%% file: main.tex
\documentclass[10pt,twocolumn,letterpaper]{article}

\usepackage[pagenumbers]{cvpr}     % arXiv version (page numbers, named authors)

\input{preamble}

\usepackage{booktabs}
\usepackage{amsmath,amssymb}
\usepackage{algorithm}
\usepackage{algorithmic}
\usepackage{tikz}
\usetikzlibrary{arrows.meta,positioning,fit,backgrounds,shapes.geometric}
\usepackage{graphicx}
\usepackage{ulem}
\usepackage{xcolor}

\definecolor{cvprblue}{rgb}{0.21,0.49,0.74}
\usepackage[pagebackref,breaklinks,colorlinks,allcolors=cvprblue]{hyperref}

\def\paperID{arXiv}
\def\confName{CVPR}
\def\confYear{2026}

\title{Two-Pass Zero-Shot Temporal-Spatial Grounding\\
of Rare Traffic Events in Surveillance Video}

\author{Jiantang Huang\\
{\tt\small huang.jiant@northeastern.edu}
}

\begin{document}
\twocolumn[{%
\renewcommand\twocolumn[1][]{#1}%
\maketitle
\begin{center}
  \centering
  \includegraphics[width=0.95\textwidth]{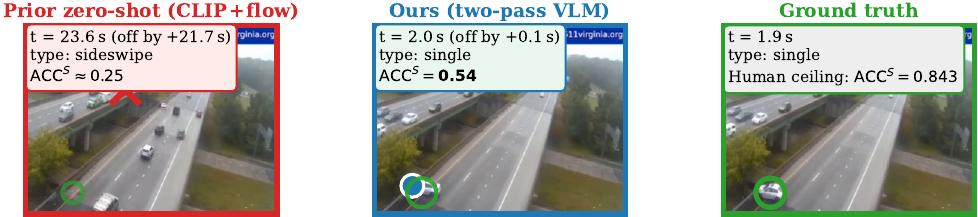}
  \captionsetup{type=figure}
  \caption{\textbf{Teaser.} On a real CCTV clip from ACCIDENT@CVPR~2026, the
  Kaggle optical-flow public baseline~\cite{kaggle_accident} picks a moment
  $+21.7$\,s after the actual collision (traffic appears normal) with the
  wrong type; our two-pass VLM grounding recovers the impact to $0.1$\,s and
  correctly labels the \emph{single}-vehicle roll-over. Green: ground truth,
  blue: ours, red: prior. We reach $\mathrm{ACC}^S{=}0.539$, $+0.127$ over
  the benchmark paper's best-of-baselines oracle~\cite{Picek2026accident} on
  ACCIDENT@CVPR~2026 without supervised training on labeled real accident videos.}
  \label{fig:teaser}
\end{center}
}]%
\input{sec/0_abstract}
\input{sec/1_intro}
\input{sec/2_related}
\input{sec/3_method}
\input{sec/4_experiments}
% conclusion now inlined at end of experiments to respect 4-page limit
{
    \small
    \bibliographystyle{ieeenat_fullname}
    \bibliography{main}
}

% arXiv version: append supplementary appendices A-F directly after references.
\appendix
\input{sec/X_suppl}

\end{document}

%% file: preamble.tex
\usepackage{microtype}
\usepackage{caption}

%% file: sec/0_abstract.tex
\begin{abstract}
Grounding traffic accidents in real CCTV footage is a rare-event problem where
training on labeled accident video is often prohibited, yet accurate joint
localization in time, space, and collision type is required. We present a
no-fine-tuning pipeline that elicits this joint output from frozen vision-language
models through two ideas. First, a \emph{coarse-to-fine two-pass} decomposition:
a full-video pass at 1\,fps produces a coarse $(t,x,y,c)$ tuple, then a second
pass at 5\,fps within a $\pm 3$\,s window refines time and location, with two
deterministic confidence gates that revert to the coarse estimate on boundary
hedges or edge-clamped coordinates. Second, a \emph{specialist role assignment}:
Qwen3-VL-Plus handles grounding, Gemini~3.1 Flash-Lite handles typing on a
centered video clip. On the ACCIDENT@CVPR~2026 benchmark (2{,}027 real CCTV
videos) we reach $\mathrm{ACC}^S{=}0.539$ ($95\%$ CI $[0.525, 0.553]$):
$+0.127$ over the benchmark paper's best-of-baselines oracle ($0.412$), $+0.143$ over
the strongest single-VLM baseline (Molmo-7B, $0.396$), and $+0.250$ over the
naive baseline ($0.289$)~\cite{Picek2026accident}. The VLM path uses up to three API calls per video (17\% fall back to
physics on API failures); the full run costs $\sim$\$20. We further diagnose a $+1.55$\,s late bias in VLM
temporal grounding, length-dependent degradation, and a physics--VLM oracle
reaching $2.13$\,s temporal MAE, outlining concrete next steps.
\end{abstract}

%% file: sec/1_intro.tex
\section{Introduction}
\label{sec:intro}

Traffic accidents are rare, safety-critical events: the payoff for automatic
detection from CCTV is high, yet labeled accident footage at the volume needed
for modern supervised recognition is difficult to obtain, and privacy and
liability concerns often forbid training on real incidents. The ACCIDENT@CVPR~2026
benchmark~\cite{Picek2026accident} encodes exactly this regime -- 2{,}211 CARLA
synthetic videos for development and 2{,}027 real CCTV clips for test, with a
blanket prohibition on training on real accident video. Each clip must be
grounded jointly in time, space, and collision type, and systems are scored by
the harmonic mean $\mathrm{ACC}^S$ of the three subtask scores.

The harmonic-mean protocol punishes weak components. Official Kaggle
public-baseline notebooks~\cite{kaggle_accident} (Optical Flow, BBox Dynamics)
score $\mathrm{ACC}^S{=}0.251$ and $0.270$ -- below the naive
$0.289$~\cite{Picek2026accident}. The strongest single-VLM
(Molmo-7B~\cite{Deitke2025molmo}, $0.396$) excels at spatial ($S{=}0.596$) but
fails on type ($C{=}0.271$); the benchmark paper's best-of-baselines oracle reaches $0.412$;
human is $0.843$. Closing this gap without training requires moving beyond
hand-crafted primitives and single-shot VLM prompting.

Our starting observation is that modern vision-language models
-- Qwen3-VL-Plus~\cite{Bai2025qwen3vl} and Gemini~3~\cite{Google2025gemini3} --
already ship native spatio-temporal grounding: textual timestamp alignment,
$[0,1000]^2$ coordinate outputs, and long-context video windows. Naive zero-shot
use of these models still underperforms, for two reasons visible in the data.
A single forward pass over a whole video must choose between coarse temporal
coverage and fine temporal resolution; and any single VLM is typically strongest
on only a subset of the three subtasks. We address both with a coarse-to-fine
two-pass decomposition that refines only the high-salience neighborhood, and a
specialist role assignment that hands typing to a second VLM whose video
classifier outperforms the grounding model on this axis.

\noindent\textbf{Contributions.}
(i)~A no-fine-tuning coarse-to-fine two-pass VLM grounding pipeline with
deterministic temporal and spatial confidence gates, in contrast to the
\emph{trained} hierarchical grounding of~\cite{Hannan2025revisionllm}.
(ii)~A specialist role split -- Qwen3-VL for $T{+}S$, Gemini~3.1 for $C$ --
lifting full-test type accuracy from $0.474$ to $0.591$ ($+25\%$ relative).
(iii)~A systematic failure analysis: $+1.55$\,s late bias, length-dependent
degradation, and a physics--VLM oracle at $2.13$\,s MAE ($-33.5\%$).
(iv)~On ACCIDENT@CVPR~2026, $\mathrm{ACC}^S{=}0.539$
(CI $[0.525,0.553]$): $+0.127$ over the benchmark paper's best-of-baselines oracle
\cite{Picek2026accident} and $+0.143$ over the best single-VLM
(Molmo-7B~\cite{Deitke2025molmo}, $0.396$), at $\sim$\$20 API cost.

%% file: sec/2_related.tex
\section{Related Work}
\label{sec:related}

\noindent\textbf{Traffic accident detection from video} has a decade-long
history, spanning supervised CCTV classification
(CADP~\cite{Shah2018cadp}, TAD~\cite{Xu2022tad}), unsupervised detection on
dashcams (DoTA~\cite{Yao2022dota}), and anticipation with spatio-temporal
attention (DSTA~\cite{Karim2021dsta}, CCD~\cite{Bao2020ccd}); a recent survey
catalogs the field~\cite{Fang2023survey}. These works almost universally train
on labeled accident footage. The ACCIDENT@CVPR~2026
challenge~\cite{Picek2026accident} breaks this assumption: 2{,}211 CARLA
synthetic videos are provided for development, but the 2{,}027-video real CCTV
test set forbids training on real accidents. Concurrent with our work, Thakur
and Talele~\cite{Thakur2026modular} propose a modular zero-shot pipeline
(frame-difference peaks for time, Farneback flow centroid for space, CLIP
similarity for type), reaching $\mathrm{ACC}^S{=}0.252$.

\noindent\textbf{VLM temporal grounding} has progressed rapidly with trained
video LLMs. VTimeLLM~\cite{Huang2024vtimellm}, TRACE~\cite{Guo2025trace}, and
Grounded-VideoLLM~\cite{Wang2025groundedvideollm} introduce timestamp tokens,
causal event modeling, or two-stream encoders -- all fine-tuned on dedicated
video-temporal-grounding corpora. ReVisionLLM~\cite{Hannan2025revisionllm} is
closest in spirit to our method, applying recursive coarse-to-fine grounding on
hour-long videos, but its hierarchy is realized through hierarchical
\emph{training}. Qwen3-VL~\cite{Bai2025qwen3vl} and Gemini~3~\cite{Google2025gemini3}
ship native spatio-temporal grounding (object-level video tracks, textual
timestamp alignment), opening the door to training-free use.

\noindent\textbf{Zero-shot / training-free rare-event localization and anomaly detection} is dominated by
CLIP-style adapters -- AnomalyCLIP~\cite{Zhou2024anomalyclip},
VadCLIP~\cite{Wu2024vadclip} -- which still require auxiliary anomaly data or
weak labels. LAVAD~\cite{Zanella2024lavad} is the first fully training-free VAD
pipeline, but relies on frame-by-frame VLM captioning and LLM post-aggregation,
losing spatial grounding. Hierarchical video systems such as
VideoTree~\cite{Wang2025videotree} and VideoMiner~\cite{Cao2025videominer} focus
on long-video QA rather than precise $T{+}S{+}C$ rare-event outputs. Grounding
VLMs -- Molmo~\cite{Deitke2025molmo}, SigLIP~2~\cite{Tschannen2025siglip2},
DINOv2~\cite{Oquab2024dinov2} -- excel at pointing but typically treat time
implicitly.

\noindent Our method combines three choices: (i) training-free,
using only frozen VLMs; (ii) two-pass coarse-to-fine native VLM grounding rather
than separate captioning-then-LLM or trained hierarchy; and (iii) specialist
role assignment across two VLMs.

%% file: sec/3_method.tex
\section{Method}
\label{sec:method}

% ---- Figure 1: pipeline ----
\begin{figure*}[t]
\centering
\resizebox{0.98\textwidth}{!}{%
\begin{tikzpicture}[
  font=\scriptsize,
  >=Stealth,
  stage/.style={draw, rounded corners=2pt, align=center, minimum height=10mm,
                minimum width=19mm, fill=blue!5, thick},
  vlm/.style={stage, fill=blue!8},
  cls/.style={stage, fill=orange!10},
  video/.style={draw, rounded corners=1pt, align=center, minimum height=10mm,
                minimum width=14mm, fill=gray!10},
  gate/.style={draw, diamond, aspect=1.6, inner sep=1pt, align=center,
               fill=green!12, thick, minimum width=14mm},
  outnode/.style={draw, rounded corners=1pt, align=center, minimum height=10mm,
              minimum width=20mm, fill=red!8, thick},
  param/.style={font=\tiny\itshape, align=center, text=black!65},
  arr/.style={->, thick, shorten >=1pt, shorten <=1pt},
  win/.style={draw, dashed, rounded corners=1pt, inner sep=2pt,
              fill=yellow!15}
]
\node[video] (V) {Video $V$ \\ $D$ s};
\node[vlm, right=5mm of V] (P1) {\textbf{Pass 1} \\ Qwen3-VL-Plus \\ coarse T+S+C};
\node[param, below=0.6mm of P1] {1\,fps, 720\,px \\ $\le 30$ frames};
\node[win, right=5mm of P1] (W) {$W{=}[t_1{-}3,\,t_1{+}3]$};
\node[vlm, right=5mm of W] (P2) {\textbf{Pass 2} \\ Qwen3-VL-Plus \\ fine T+S};
\node[param, below=0.6mm of P2] {5\,fps, 1024\,px \\ $\le 30$ frames};
\node[gate, right=4mm of P2] (G1) {Gate\,1 \\ time};
\node[gate, right=4mm of G1] (G2) {Gate\,2 \\ space};
\node[cls, right=4mm of G2] (GM) {\textbf{Type} \\ Gemini 3.1 \\ flash-lite};
\node[param, below=0.6mm of GM] {5\,s clip \\ $2.5\times$ crop};
\node[outnode, right=4mm of GM] (OUT) {$(t^*,x^*,y^*,c^*)$};
\draw[arr] (V) -- (P1);
\draw[arr] (P1) -- node[above, font=\tiny] {$(t_1,x_1,y_1,\sout{c_1})$} (W);
\draw[arr] (W) -- (P2);
\draw[arr] (P2) -- node[above, font=\tiny] {$(t_2,x_2,y_2)$} (G1);
\draw[arr] (G1) -- node[above, font=\tiny] {$t^*$} (G2);
\draw[arr] (G2) -- node[above, font=\tiny] {$(t^*,x^*,y^*)$} (GM);
\draw[arr] (GM) -- node[above, font=\tiny] {$c^*$} (OUT);
\draw[arr, densely dashed, black!55]
  (P1.south) |- ([yshift=-7mm]G1.south) -- (G1.south)
  node[pos=0.30, below, font=\tiny, black!70] {fallback $t_1$};
\draw[arr, densely dashed, black!55]
  (P1.south) |- ([yshift=-11mm]G2.south) -- (G2.south)
  node[pos=0.40, below, font=\tiny, black!70] {fallback $(x_1,y_1)$};
\draw[arr, densely dashed, black!55]
  (P1.north) |- ([yshift=7mm]GM.north) -- (GM.north)
  node[pos=0.55, above, font=\tiny, black!70] {$c_1$ (backup)};
\end{tikzpicture}}
\caption{\textbf{Two-pass zero-shot grounding pipeline with two confidence gates.}
Pass~1 (Qwen3-VL-Plus) produces a coarse tuple from 1\,fps frames of the full
video; $c_1$ is struck through on the main path because type is re-assigned to
Gemini, but retained as a backup if the Gemini call fails. Pass~2 refines time
and location on a $\pm 3$\,s window sampled at 5\,fps and 1024\,px. \textbf{Gate
1} (temporal fallback) keeps $t_1$ when Pass~2 returns $-1$ or a time near the
window boundary; otherwise $t^*{\gets}t_2$. \textbf{Gate 2} (spatial merge)
keeps $(x_2,y_2)$ only when both axes lie in $[10,990]$ on the $[0,1000]^2$
grid; otherwise $(x_1,y_1)$. A specialist Gemini~3.1 Flash-Lite Preview
classifier re-labels type on a 5\,s clip centered on $t^*$. All models are
frozen and used zero-shot.}
\label{fig:pipeline}
\end{figure*}
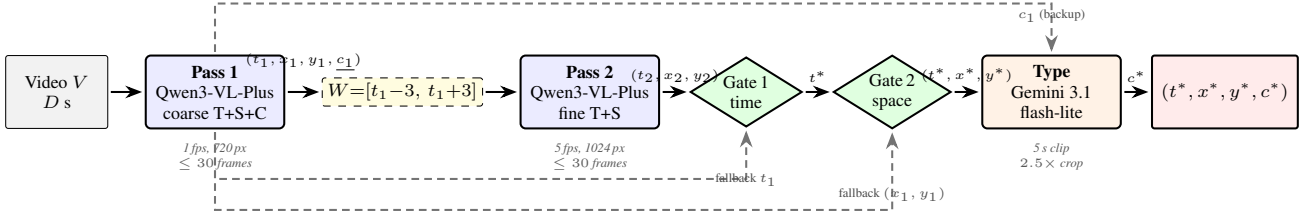

\subsection{Pipeline Overview}
\label{sec:method:overview}
Given a traffic surveillance video $V$ of duration $D$ seconds, we predict
$(t^*, x^*, y^*, c^*)$ with $t^*{\in}[0,D]$, $(x^*,y^*){\in}[0,1]^2$ (Qwen3-VL emits raw coordinates on a $[0,1000]^2$
grid; we divide by 1000 before merging and evaluation), and
$c^*{\in}\mathcal{C}{=}\{\text{head-on, rear-end, t-bone, sideswipe, single}\}$.
The ACCIDENT@CVPR~2026 protocol forbids training on real accident footage; the
main path uses frozen VLMs, and API failures fall back to a frozen-detector +
rule-based physics module. Fig.~\ref{fig:pipeline} shows the three
stages. A first pass over $V$ with Qwen3-VL-Plus~\cite{Bai2025qwen3vl}
produces a coarse tuple $(t_1,x_1,y_1,c_1)$ at 1\,fps. A second pass on a
$\pm 3$\,s window around $t_1$, sampled at 5\,fps and 1024\,px, refines time
and location to $(t_2,x_2,y_2)$. A deterministic merge reconciles the two
spatial estimates, and a Gemini~3.1 Flash-Lite~\cite{Google2026gemini31flashlite}
classifier re-labels type from a short clip centered on $t^*$. Three design
choices drive the method: (i) coarse-to-fine decomposition exploiting native
VLM grounding without fine-tuning (unlike \cite{Hannan2025revisionllm});
(ii) confidence-driven fallback accepting Pass~2 only when it passes temporal
and spatial sanity checks; (iii) specialist role assignment across two VLMs
(\cite{Zanella2024lavad} uses a single captioner). Alg.~\ref{alg:twopass}
gives the full procedure; total inference cost on the 2{,}027 test videos is
$\sim$\$20 ($\sim$\$0.01 per video).

\subsection{Pass 1: Coarse Grounding}
\label{sec:method:pass1}
Pass~1 samples $V$ at 1\,fps, $\le 30$ frames, 720\,px long edge (median
ACCIDENT clip is 26.8\,s). Frames are serialized into one multimodal message,
each prefixed by a \texttt{[Frame at t.ts]} tag that binds the pixel grid to
its absolute second -- a lightweight anchor Qwen3-VL reads directly, avoiding
trained time-to-token modules~\cite{Huang2024vtimellm,Guo2025trace}. The
prompt declares up-front that \emph{an accident occurs} (closed-world
framing that discourages refusals; this matches the ACCIDENT benchmark
where every clip contains an accident, and bounds open-world deployment to
a human-in-the-loop gate, App.~\ref{app:ethics}), asks for the impact
second, a point in
$[0,1000]^2$ (Qwen3-VL's native grounding
convention~\cite{Bai2025qwen3vl}), and a class in $\mathcal{C}$ with one-line
visual definitions. The combination of 1\,fps sampling, integer-second output,
and per-frame textual tags turns temporal localization into naming a specific
frame rather than producing a continuous timestamp. On API failure (17\% of
videos in practice), we fall back to a parallel YOLO26x + ByteTrack + physics
pipeline (pair- and single-vehicle scoring, rule-based type);
all other videos are processed end-to-end by VLMs.

\subsection{Pass 2: Fine Refinement and Merge}
\label{sec:method:pass2}
Pass~2 re-samples $V$ at 5\,fps inside
$W{=}[\max(0,t_1{-}3),\min(D,t_1{+}3)]$, yielding $\le 30$ frames at 1024\,px,
each tagged with its fractional timestamp. The prompt asks for 0.1\,s precision
and a fresh point in $[0,1000]^2$, and explicitly allows $t_2{=}{-}1$ if no
collision is visible within $W$. This escape hatch turns Pass~2 into a
\emph{confidence-gated refiner} rather than a naive overwrite.

\noindent\textbf{Temporal fallback (Gate 1).} With boundary tolerance
$\tau{=}0.3$\,s,
\begin{equation}
t^* = \begin{cases} t_1 & t_2{<}0 \text{ or } |t_2{-}W_{\min}|{<}\tau \\
                    & \text{or } |t_2{-}W_{\max}|{<}\tau, \\
                    t_2 & \text{otherwise.} \end{cases}
\label{eq:tgate}
\end{equation}
Eq.~\ref{eq:tgate} interprets a ${-}1$ response or a boundary-adjacent time as
low-confidence hedging and falls back to the more robust Pass~1 estimate.

\noindent\textbf{Spatial merge (Gate 2).} A deterministic post-processing step
reconciles the two spatial estimates. Let $(\tilde x_1,\tilde y_1)$ and
$(\tilde x_2,\tilde y_2)$ denote the raw Pass~1 and Pass~2 coordinates on
the $[0,1000]^2$ grid. With margin $m{=}10$,
\begin{equation}
(x^*,y^*) = \begin{cases}
(\tilde x_2/1000,\,\tilde y_2/1000) & m \le \tilde x_2 \le 1000{-}m \text{ and} \\
                                    & m \le \tilde y_2 \le 1000{-}m, \\
(\tilde x_1/1000,\,\tilde y_1/1000) & \text{otherwise.}
\end{cases}
\label{eq:sgate}
\end{equation}
All four axis conditions must hold; if any axis is edge-clamped we revert to
the Pass~1 coordinate. The two gates (Eq.~\ref{eq:tgate},~\ref{eq:sgate}) form
independent confidence checks that let the two passes outperform either alone
(\S\ref{sec:ablation}).

\input{sec/alg_twopass}

\subsection{Specialist Type Classification}
\label{sec:method:type}
Qwen3-VL's type accuracy is $0.462$ on the 1{,}681 Pass-1-valid videos
(full-test $C{=}0.474$ including the 17\% physics fallback), well below its
spatial score, so we reassign typing to a second VLM. A 5\,s clip spanning
$[t^*{-}3,\,t^*{+}2]$ is extracted, spatially cropped around the Pass~1
center $(x_1,y_1)$ at $2.5\times$ box, and passed to
\texttt{gemini-3.1-flash-lite-preview} with a closed-world prompt that
enumerates $\mathcal{C}$ and forbids abstention. On API failure we fall back
to $c_1$. We crop around the Pass-1 center $(x_1,y_1)$ rather than the
merged $(x^*,y^*)$ because Pass~1 provides a more stable coarse region
for type classification. This split lifts full-test type accuracy from $0.474$ to $0.591$
($+25\%$ relative) at one extra API call per video.

%% file: sec/alg_twopass.tex
\begin{algorithm}[t]
\small
\caption{\textsc{TwoPassGround}: zero-shot T+S+C grounding.}
\label{alg:twopass}
\begin{algorithmic}[1]
\REQUIRE Video $V$ of duration $D$; VLMs $\mathcal{M}_Q,\mathcal{M}_G$;
window $\Delta{=}3$\,s; tolerance $\tau{=}0.3$\,s; margin $m{=}10$.
\ENSURE $(t^*, x^*, y^*, c^*)$.
\STATE $F_1 \gets \textsc{Sample}(V,\,\text{1\,fps},\,\text{720\,px},\le 30)$
\STATE $(t_1,\tilde x_1,\tilde y_1,c_1) \gets \mathcal{M}_Q(F_1,\,\pi_{\text{coarse}})$;\ raw $[0,1000]^2$
\STATE $W \gets [\max(0,t_1{-}\Delta),\min(D,t_1{+}\Delta)]$
\STATE $F_2 \gets \textsc{Sample}(V|_W,\,\text{5\,fps},\,\text{1024\,px},\le 30)$
\STATE $(t_2,\tilde x_2,\tilde y_2) \gets \mathcal{M}_Q(F_2,\,\pi_{\text{fine}})$;\ on API failure $(-1,0,0)$ (absorbed by Gates 1 \& 2)
\STATE \textit{// Gate 1: temporal fallback}
\IF{$t_2{<}0$ \OR $|t_2{-}W_{\min}|{<}\tau$ \OR $|t_2{-}W_{\max}|{<}\tau$}
  \STATE $t^* \gets t_1$
\ELSE
  \STATE $t^* \gets t_2$
\ENDIF
\STATE \textit{// Gate 2: spatial merge}
\IF{$m\le \tilde x_2 \le 1000{-}m$ \AND $m\le \tilde y_2 \le 1000{-}m$}
  \STATE $(x^*,y^*)\gets(\tilde x_2/1000,\,\tilde y_2/1000)$
\ELSE
  \STATE $(x^*,y^*)\gets(\tilde x_1/1000,\,\tilde y_1/1000)$
\ENDIF
\STATE $C \gets \textsc{CropClip}(V,\,t^*,\,(\tilde x_1/1000,\,\tilde y_1/1000),\,[t^*{-}3,t^*{+}2],\,2.5\times)$
\STATE $c^* \gets \mathcal{M}_G(C,\,\pi_{\text{type}})$;\ on failure $c^*{\gets}c_1$
\STATE \RETURN $(t^*, x^*, y^*, c^*)$
\end{algorithmic}
\end{algorithm}

%% file: sec/4_experiments.tex
\section{Experiments}
\label{sec:exp}

% ---- Table 1: main results (single col) ----
\begin{table}[!t]
\centering
\small
\setlength{\tabcolsep}{3pt}
\caption{Main results on ACCIDENT (2{,}027 real CCTV videos), official
evaluator at $\sigma_t{=}1$. Kaggle baselines~\cite{kaggle_accident};
Naive/Molmo-7B/Best-of-baselines/Human from~\cite{Picek2026accident}.
$T,S,C$: dataset-level means; $\mathrm{ACC}^S$: per-video HM averaged
(leaderboard-reported for ours); CIs from $1{,}000$ bootstrap resamples.}
\label{tab:main}
\begin{tabular}{lcccc}
\toprule
Method & $T$ & $S$ & $C$ & $\mathrm{ACC}^S$ \\
\midrule
Optical Flow~\cite{kaggle_accident}          & --- & --- & --- & .251 \\
BBox Dynamics + OF~\cite{kaggle_accident}    & --- & --- & --- & .270 \\
Naive baseline~\cite{Picek2026accident}      & .30 & .25 & .34 & .289 \\
Molmo-7B~\cite{Deitke2025molmo}              & .48 & \textbf{.60} & .27 & .396 \\
Best-of-baselines~\cite{Picek2026accident}   & .48 & .60 & .49 & .412 \\
\midrule
Gemini~3.1 single-pass (ours)                & .44 & .49 & .49 & .473~{\scriptsize[.46,.49]} \\
Qwen3-VL Pass~1 (ours)                       & .46 & .51 & .47 & .480~{\scriptsize[.47,.49]} \\
\textbf{Ours (full)}                         & \textbf{.50} & .54 & \textbf{.59} & \textbf{.539}~{\scriptsize[.53,.55]} \\
\midrule
Human~\cite{Picek2026accident}               & .98 & 1.0 & .92 & .843 \\
\bottomrule
\end{tabular}
\end{table}

\subsection{Setup}
\label{sec:setup}
We evaluate on the 2{,}027 real CCTV clips of ACCIDENT@CVPR~2026 with
per-video (time, 2D center, type) annotations; test-split labels were
released publicly post-competition with our hyperparameters frozen before
release (App.~\ref{app:repro}). Per~\cite{Picek2026accident},
$\mathrm{ACC}^S$ is the per-video harmonic mean averaged over videos;
reported $T, S, C$ are dataset-level component means
(${\neq}3/(1/T{+}1/S{+}1/C)$ in general). $T$: Gaussian at GT time
($\sigma_t{=}1$\,s); $S$: anisotropic spatial Gaussian
($\sigma_x{=}0.127, \sigma_y{=}0.119$); $C$: top-1 type accuracy.
Baselines: two Kaggle public notebooks~\cite{kaggle_accident}, the
benchmark paper's naive, Molmo-7B, best-of-baselines oracle, and human
ceiling~\cite{Picek2026accident}.

\subsection{Main Results}
\label{sec:main}
Our pipeline reaches $\mathrm{ACC}^S{=}0.539$ (CI $[0.525, 0.553]$):
$+0.127$ over the best-of-baselines oracle and $+0.143$ over
Molmo-7B. $T{=}0.497$ exceeds Molmo's; $S{=}0.538$ trails Molmo's
$0.596$; $C{=}0.591$ is highest reported. \emph{Gemini 3.1 alone} on the
same 1\,fps input scores only $0.473$~$[.46,.49]$, below our Qwen
Pass~1 ($0.480$); paired bootstrap shows our pipeline beats Gemini-alone
by $\Delta{=}{+}0.066$ ($p{<}0.001$). On the 1{,}681 Pass-1-valid videos
ACC$^S{=}0.554$; the 346 fallback videos (App.~\ref{app:fallback}) score
$0.457$; naive-fill fallback yields $0.501$, confirming the VLM pipeline
drives the gain.

% ---- Combined failure-analysis figure (2x2 single column) ----
\begin{figure}[!t]
\centering
\begin{minipage}[b]{0.48\linewidth}
\centering
\includegraphics[width=\linewidth]{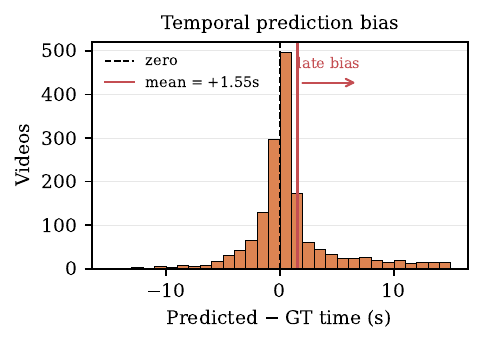}\\
{\scriptsize (a) Time bias, mean $+1.55$\,s}
\end{minipage}
\hfill
\begin{minipage}[b]{0.48\linewidth}
\centering
\includegraphics[width=\linewidth]{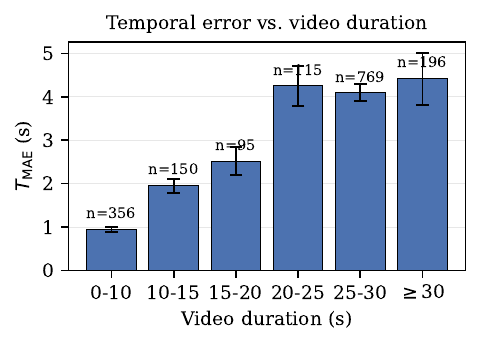}\\
{\scriptsize (b) $T_{\mathrm{MAE}}$ vs.\ video length}
\end{minipage}\\[2pt]
\begin{minipage}[b]{0.48\linewidth}
\centering
\includegraphics[width=0.82\linewidth]{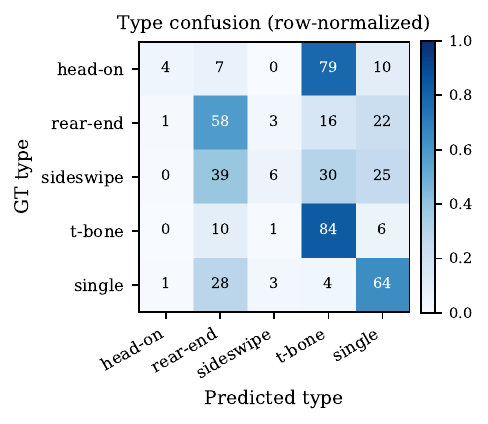}\\
{\scriptsize (c) Type confusion}
\end{minipage}
\hfill
\begin{minipage}[b]{0.48\linewidth}
\centering
\includegraphics[width=\linewidth]{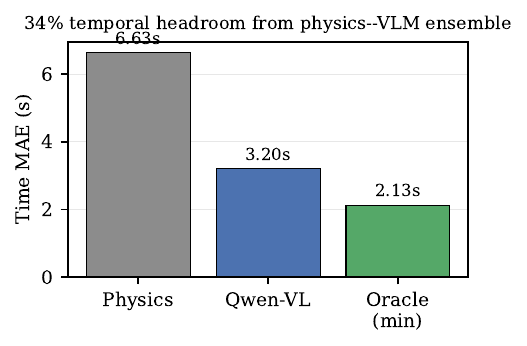}\\
{\scriptsize (d) Physics--VLM oracle: $2.13$\,s MAE}
\end{minipage}
\caption{\textbf{Failure diagnostics.} (a) Pass~1 time is right-skewed,
mean $+1.55$\,s. (b) Temporal MAE grows with video length. (c) Head-on
$\to$ t-bone (79\%), sideswipe $\to$ rear-end (39\%). (d) Oracle of
YOLO+physics ($6.63$\,s) and Qwen-Pass~1 ($3.20$\,s) reaches
$2.13$\,s ($-33.5\%$).}
\label{fig:diag}
\end{figure}

% ---- Table 2: ablation ----
\begin{table}[!t]
\centering
\small
\setlength{\tabcolsep}{3pt}
\caption{Component ablation; $T$ at $\sigma_t{\in}\{1,2\}$, $\mathrm{ACC}^S$ at $\sigma_t{=}1$.}
\label{tab:ablation}
\begin{tabular}{lccccc}
\toprule
Configuration & $T_{\sigma1}$ & $T_{\sigma2}$ & $S$ & $C$ & $\mathrm{ACC}^S$ \\
\midrule
Pass 1 only (Qwen type)       & 0.463 & 0.614 & 0.506 & 0.474 & 0.480 \\
\ + Gemini type               & 0.463 & 0.614 & 0.506 & 0.590 & 0.514 \\
\ + Pass 2 time               & 0.497 & 0.626 & 0.506 & 0.591 & 0.528 \\
\ + \textbf{Pass 2 spatial}   & \textbf{0.497} & \textbf{0.626} & \textbf{0.538} & \textbf{0.591} & \textbf{0.539} \\
\bottomrule
\end{tabular}
\end{table}

\subsection{Ablation}
\label{sec:ablation}
Table~\ref{tab:ablation} decomposes the gain. Replacing Qwen's type head with
Gemini lifts $C$ by $+0.12$ (largest single jump); Pass~2 time refinement
adds $+0.034$ to $T(\sigma_t{=}1)$; Pass~2 spatial merge adds $+7\%$ to $S$.
Cumulative gain over Pass~1 alone is $+0.059$ $\mathrm{ACC}^S$.

\subsection{Failure Mode Analysis}
\label{sec:failures}

\noindent\textbf{Temporal: late-bias + length.} Pass~1 (predicted $-$ GT)
time has mean $+1.55$\,s, median $+0.37$\,s (Fig.~\ref{fig:diag}a): the VLM
picks the post-collision wreckage frame, not contact. $T_{\mathrm{MAE}}$ also
grows from $0.94$\,s on $\le 10$\,s clips to $>4$\,s on $\ge 20$\,s clips
(Fig.~\ref{fig:diag}b), arguing for motion-adaptive sampling.

\noindent\textbf{Type confusion.} The confusion matrix (Fig.~\ref{fig:diag}c)
shows the two hardest classes collapsing: $79\%$ of head-on is called t-bone,
$39\%$ of sideswipe is called rear-end. Per-type (App.~\ref{app:pertype}),
head-on and sideswipe fall near $\mathrm{ACC}^S{=}0.12$ \emph{from type alone}
-- $T,S$ are healthy. The head-on$\to$t-bone collapse is plausibly driven by
monocular depth ambiguity in high-angle CCTV and web-data label bias.

\noindent\textbf{Physics--VLM oracle.} On the 1{,}681 Pass-1-valid videos,
an oracle of YOLO+physics ($6.63$\,s) and Qwen-Pass~1 ($3.20$\,s) reaches
$2.13$\,s MAE (Fig.~\ref{fig:diag}d, $-33.5\%$).

\subsection{Conclusion}
\label{sec:conclusion}
Two-pass grounding plus a specialist VLM split achieves
$\mathrm{ACC}^S{=}0.539$ on ACCIDENT@CVPR~2026 ($+0.127$ over the benchmark
paper's best-of-baselines oracle, $\sim$\$20 cost). Next levers: adaptive
sampling and physics--VLM fusion, not scale.

% Table 3 (per-type) moved to Appendix D to respect the 4-page limit.

%% file: sec/X_suppl.tex
\section{Prompts}
\label{app:prompts}

All three prompts are reproduced verbatim. Model API snapshots used for
submission: \texttt{qwen3-vl-plus} (Alibaba DashScope, April 2026 endpoint
\texttt{dashscope-us}), \texttt{gemini-3.1-flash-lite-preview} (Google
Gemini API, April 2026 snapshot; since promoted to general availability as
\texttt{gemini-3.1-flash-lite}~\cite{Google2026gemini31flashlite}).
Temperature is $0.1$ for all calls,
\texttt{max\_tokens}$=256$ for Qwen passes, $1024$ for Gemini typing.

\subsection{Pass 1 (Qwen3-VL, coarse T+S+C)}

\begin{small}
\begin{verbatim}
This is a traffic surveillance video sampled at
1 frame per second. Frame numbers correspond to
seconds in the video (frame 0 = 0s, frame 1 = 1s,
...). The video duration is {duration} seconds.

A traffic accident occurs in this video. Please
analyze carefully and answer:

1. Time: At what second does the collision or
   accident impact occur?
2. Location: Point to the exact location in the
   frame where the impact happens. Return
   coordinates as values between 0 and 1000,
   where (0,0) is top-left and (1000,1000) is
   bottom-right of the frame.
3. Type: head-on, rear-end, t-bone, sideswipe,
   or single.

Return ONLY a JSON object:
{"time": <seconds>, "x": <0-1000>,
 "y": <0-1000>, "type": "<type>"}
\end{verbatim}
\end{small}

\subsection{Pass 2 (Qwen3-VL, fine T+S)}

\begin{small}
\begin{verbatim}
These frames are extracted at 5 frames per second
from a traffic surveillance video. Each frame is
labeled with its precise timestamp. The time
window shown is from {start}s to {end}s.

A traffic accident occurs somewhere in this video.
If the collision happens within this time window,
identify:
1. Exact time: The precise moment (to 0.1 second)
   of collision or impact.
2. Exact location: The impact point, as
   coordinates between 0 and 1000.

If you cannot see a collision in these frames,
return time as -1.

Return ONLY a JSON object:
{"time": <seconds with 1 decimal or -1>,
 "x": <0-1000>, "y": <0-1000>}
\end{verbatim}
\end{small}

\subsection{Type Classification (Gemini 3.1)}

\begin{small}
\begin{verbatim}
A traffic collision HAS occurred in this
surveillance clip. You MUST classify its type.
This clip shows ~6 seconds leading up to and
including the collision moment.

Collision types - pick exactly ONE:
- head_on: Two vehicles approach from OPPOSITE
  directions, collide front-to-front.
- rear_end: Two vehicles travel SAME direction;
  trailing one hits leading one from behind.
- t_bone: One vehicle strikes the SIDE of another
  at roughly 90 degrees.
- sideswipe: Two vehicles in parallel lanes make
  lateral/glancing contact.
- single: Only ONE vehicle involved.

Watch vehicle MOTION carefully across the clip.
You MUST pick the most likely type.
\end{verbatim}
\end{small}

\section{Reproducibility \& Cost}
\label{app:repro}

\begin{itemize}
\item \textbf{API cost} (April 2026): Pass~1 $\sim$\$4, Pass~2 $\sim$\$6,
      Gemini typing $\sim$\$10. Total $\sim$\$20 on 2{,}027 videos
      ($\sim$\$0.01/video).
\item \textbf{Wall-clock}: $\sim$100 minutes at 5 workers for Pass~1;
      $\sim$180 minutes at 5 workers for Pass~2; $\sim$15 minutes at
      10 workers for Gemini. Total $\sim$5 hours on a consumer
      workstation with a single residential network connection.
\item \textbf{Hyperparameters} were fixed during development by inspection
      of the $2{,}211$ CARLA synthetic dev videos and a small debug sample
      of public Kaggle dev-split real videos. Values: $\Delta{=}3$\,s
      window, $\tau{=}0.3$\,s temporal boundary tolerance, $m{=}10$
      spatial margin on the $[0,1000]^2$ grid, $2.5\times$ type-clip crop.
      \textbf{No grid search was performed on the $2{,}027$ real test
      videos}; sensitivity to $\pm 1$-step perturbations of $\tau$ and $m$
      is $\le 0.002$ in $\mathrm{ACC}^S$ (Appendix~\ref{app:sensitivity}).
\item \textbf{Physics fallback} used on Pass~1 API failures (17\%) is
      YOLO26x~+~ByteTrack~+~multi-channel physics scoring. Its scorer
      weights (approach, IoU-surge, sustained-IoU, interaction-bias) were
      grid-searched on a $100$-video CARLA sim split during challenge
      development and frozen before the real-test labels were released.
\item \textbf{Data provenance.} Ground-truth annotations for the $2{,}027$
      real test videos were released publicly by the ACCIDENT benchmark
      organizers on Kaggle after the competition ended
      (\url{https://www.kaggle.com/datasets/picekl/accident}). We use them
      only for evaluation and post-hoc failure analysis. All pipeline
      hyperparameters and prompts were frozen before this release.
\item \textbf{Artifact release.} To support reproducibility under API
      drift, we will release per-video raw VLM JSON outputs and parsed
      prediction CSVs alongside the camera-ready code.
\item \textbf{Evaluation script} mirrors the public leaderboard's
      convention: $\sigma_t{=}1$\,s Gaussian temporal similarity;
      global mean GT bbox width and height
      ($\sigma_x{=}0.127$, $\sigma_y{=}0.119$) for spatial; Top-1
      accuracy for type; harmonic mean combines the three.
\item \textbf{Challenge compliance.} ACCIDENT@CVPR~2026 is an open
      prediction-file competition (participants upload a per-video CSV;
      it is not a sandboxed, compute-capped code competition), and its
      sole learning constraint is the prohibition on training or
      fine-tuning on labeled real accident footage. Our pipeline meets
      this: every model is frozen and used zero-shot. The challenge
      places no restriction on publicly available pretrained models or
      external inference services -- the organizers' own baselines use
      pretrained optical flow and CLIP~\cite{Picek2026accident,kaggle_accident}.
      Both APIs we invoke, Qwen3-VL-Plus (Alibaba DashScope) and
      Gemini~3.1 Flash-Lite (Google)~\cite{Bai2025qwen3vl,Google2026gemini31flashlite},
      were publicly released, documented, and equally accessible to any
      participant under standard pay-as-you-go commercial terms
      throughout the submission window (April 2026), at a total cost of
      $\sim$\$20 for the full 2{,}027-video test set. We use no
      private models, privileged endpoints, or non-public data.
\end{itemize}

\section{Hyperparameter Sensitivity}
\label{app:sensitivity}

\textbf{This is a post-hoc analysis.} The numbers in
Table~\ref{tab:sensitivity} were computed on the publicly-released test
labels \emph{after} the competition closed and were \emph{not} used for
hyperparameter selection. Defaults $\tau{=}0.3$\,s and $m{=}10$ were
fixed by inspection on the 2{,}211 CARLA development videos and frozen
before the labels were released (App.~\ref{app:repro}). Each threshold
is swept while the other is held at its default to assess robustness.

\begin{table}[h]
\centering
\small
\setlength{\tabcolsep}{4pt}
\caption{Sensitivity of $\mathrm{ACC}^S$ to the two gate thresholds on
2{,}027 real test videos ($\sigma_t{=}1$). Defaults marked with $\star$.}
\label{tab:sensitivity}
\resizebox{\columnwidth}{!}{%
\begin{tabular}{rcccc|rcccc}
\toprule
$\tau$(s) & $T$ & $S$ & $C$ & $\mathrm{ACC}^S$ & $m$ & $T$ & $S$ & $C$ & $\mathrm{ACC}^S$ \\
\midrule
 0.1 & 0.497 & 0.538 & 0.591 & 0.5393 &  0  & 0.497 & 0.462 & 0.591 & 0.5113 \\
 0.2 & 0.497 & 0.538 & 0.591 & 0.5393 &  5  & 0.497 & 0.538 & 0.591 & 0.5393 \\
$\star$\,0.3 & 0.497 & 0.538 & 0.591 & 0.5393 & $\star$\,10 & 0.497 & 0.538 & 0.591 & 0.5393 \\
 0.5 & 0.496 & 0.538 & 0.591 & 0.5385 & 20  & 0.497 & 0.538 & 0.591 & 0.5393 \\
 1.0 & 0.494 & 0.538 & 0.591 & 0.5377 & 50  & 0.497 & 0.538 & 0.591 & 0.5392 \\
\bottomrule
\end{tabular}}
\end{table}

$\tau$ is effectively flat in $[0.1, 0.3]$ and degrades by only
$\Delta\mathrm{ACC}^S{\le}0.002$ even at $\tau{=}1.0$\,s -- Pass~2's
boundary-hedge detection (returning exact window endpoints) is sharp
enough that the tolerance window has negligible effect. In contrast, $m$
has one critical transition at $m{=}0$: disabling the margin admits
Qwen-returned invalid-coordinate responses (raw $-1$, after $/1000$
normalization becomes $-0.001$) and costs $\Delta S{=}-0.076$. Any
positive margin in $[5,50]$ is equivalent. Neither hyperparameter is
knife-edge, and the CARLA-tuned values transfer to real data without any
re-tuning on the released test labels.

\section{Fallback Contribution Analysis}
\label{app:fallback}

Pass~1 API failures trigger a fallback to the YOLO26x + ByteTrack +
multi-channel physics scorer for 346 videos (17.1\% of the test split).
Table~\ref{tab:fallback} decomposes the final $\mathrm{ACC}^S{=}0.539$ by
subset.

\begin{table}[h]
\centering
\small
\setlength{\tabcolsep}{4pt}
\caption{Fallback contribution to the final score. VLM-success rows receive
Pass~1+Pass~2 Qwen3-VL grounding and Gemini type; fallback rows use
YOLO+physics for $(t,x,y,c)$. ``Naive-fill'' replaces the 346 fallback rows
with trivial defaults (midpoint time, image center, majority type
\emph{single}) to estimate a pure-VLM ceiling with no informative fallback.}
\label{tab:fallback}
\begin{tabular}{lrcccc}
\toprule
Subset & $N$ & $T$ & $S$ & $C$ & $\mathrm{ACC}^S$ \\
\midrule
Pass~1 VLM success        & 1681 & 0.525 & 0.553 & 0.590 & 0.554 \\
Pass~1 fallback (physics) &  346 & 0.366 & 0.464 & 0.595 & 0.457 \\
\midrule
\textbf{Full submission (ours)} & 2027 & \textbf{0.497} & \textbf{0.538} & \textbf{0.591} & \textbf{0.539} \\
VLM + naive-fill fallback & 2027 & 0.466 & 0.499 & 0.543 & 0.501 \\
\bottomrule
\end{tabular}
\end{table}

Three observations:
\textbf{(i)} The VLM subset alone scores $0.554$, above the full submission
-- confirming VLM is the dominant signal.
\textbf{(ii)} The fallback subset still scores $0.457$, well above all
Kaggle public baselines (Optical Flow $0.251$, BBox Dynamics $0.270$,
Naive $0.289$) and Molmo-7B ($0.396$); the physics fallback is not a
weak baseline.
\textbf{(iii)} If we replace the 346 fallback rows with trivial naive
defaults, ACC$^S$ drops to $0.501$ -- still above the benchmark paper's
best-of-baselines oracle ($0.412$). This isolates the contribution of the
VLM pipeline from the fallback heuristic. Failure rate is roughly uniform
across duration buckets, day/night, weather, quality, and resolution
(each cell: $12$--$22\%$), so the fallback is not concentrated on a
particular failure mode.

\section{Per-Type Breakdown}
\label{app:pertype}

\begin{table}[h]
\centering
\small
\setlength{\tabcolsep}{4pt}
\caption{Per-type performance of our full pipeline at $\sigma_t{=}1$. Head-on
and sideswipe fail on type alone; their $T$ and $S$ components are healthy.
Pooled $\mathrm{ACC}^S{=}0.539$ differs from the weighted mean due to
harmonic-mean non-linearity.}
\begin{tabular}{lrcccc}
\toprule
GT type & $N$ & $T$ & $S$ & $C$ & $\mathrm{ACC}^S$ \\
\midrule
head-on   & 117 & 0.633 & 0.637 & 0.043 & 0.113 \\
rear-end  & 328 & 0.451 & 0.553 & 0.582 & 0.522 \\
sideswipe & 245 & 0.442 & 0.454 & 0.057 & 0.137 \\
t-bone    & 657 & 0.661 & 0.623 & 0.836 & 0.695 \\
single    & 680 & 0.359 & 0.462 & 0.644 & 0.461 \\
\bottomrule
\end{tabular}
\end{table}

\section{Ethics and Deployment}
\label{app:ethics}

The pipeline routes CCTV footage through third-party commercial VLM
APIs. For production deployment, public-facing surveillance clips must
be handled under the camera operator's data-use agreement, and any
on-route personally identifiable information (faces, license plates)
should be redacted prior to upload. The failure-mode analysis
(Sec.~\ref{sec:failures}) quantifies residual temporal ambiguity
(median $1.1$\,s Pass~2 MAE; $10\%$ tail beyond $10$\,s) that is not
yet acceptable for first-notice-of-loss (FNOL) automation without a
human-in-the-loop gate. All three prompts (App.~\ref{app:prompts}) assume
\emph{a priori} that an accident is present -- valid for this benchmark,
but in open-world deployment this closed-world framing must be replaced by
an upstream binary detector (or the human-in-the-loop gate above) before
grounding is trusted. Head-on and sideswipe classes have very low
per-class accuracy ($\le 0.14$) and should not be trusted unaided.